\pdfoutput=1

\documentclass[11pt]{article}

\usepackage{iclr2022_conference,times}
\iclrfinalcopy 

\usepackage{times}
\usepackage{latexsym}

\usepackage[utf8]{inputenc} 
\usepackage[T1]{fontenc}    
\usepackage{hyperref}       
\usepackage{url}            
\usepackage{booktabs}       
\usepackage{amsfonts}       
\usepackage{nicefrac}       
\usepackage{microtype}      
\usepackage{xcolor}         
\usepackage{bm,bm,bbm,dsfont}
\usepackage{comment}
\usepackage{pgfplots}
\usepackage{arydshln}
\usepackage{pifont}
\usepackage{amsfonts}
\usepackage{multicol,multirow}
\usepackage{amsmath, amssymb, amsthm}
\usepackage[linesnumbered,ruled,vlined]{algorithm2e}

\SetKwInput{KwInput}{Input}                
\SetKwInput{KwOutput}{Output}              

\usepackage{bm,bbm,dsfont}
\pgfplotsset{width=7cm,compat=1.13}

\usepackage[T1]{fontenc}
\usepackage[utf8]{inputenc}

\usepackage{microtype}

\usepackage{microtype}
\definecolor{ao}{rgb}{0.0, 0.5, 0.0}
\definecolor{asparagus}{rgb}{0.53, 0.66, 0.42}
\definecolor{amber}{rgb}{1.0, 0.49, 0.0}
\definecolor{alizarin}{rgb}{0.82, 0.1, 0.26}
\definecolor{applegreen}{rgb}{0.55, 0.71, 0.0}
\definecolor{amethyst}{rgb}{0.6, 0.4, 0.8}
\definecolor{auburn}{rgb}{0.43, 0.21, 0.1}

\usepackage{amsmath,amsfonts,bm}

\def\eqref#1{equation~\ref{#1}}

\def\1{\bm{1}}

\DeclareMathAlphabet{\mathsfit}{\encodingdefault}{\sfdefault}{m}{sl}
\SetMathAlphabet{\mathsfit}{bold}{\encodingdefault}{\sfdefault}{bx}{n}

\DeclareMathOperator*{\argmax}{arg\,max}
\DeclareMathOperator*{\argmin}{arg\,min}

\title{A General Framework for Defending against Backdoor Attacks via Influence Graph}

\author{Xiaofei Sun$^{1}$, Jiwei Li$^{1,2}$, 
Xiaoya Li$^{1}$, Ziyao Wang$^{3}$ \\
\bf Tianwei Zhang$^{4}$, Han Qiu$^{5}$, Fei Wu$^{2}$, Chun Fan$^{6}$\\
$^{1}$Shannon.AI,~$^{2}$Zhejiang University,
~$^{3}$University of Leeds,
~$^{4}$Nanyang Technological University\\
~$^{5}$Tsinghua University, ~$^{6}$ Peking University\\
\{xiaofei\_sun, jiwei\_li, xiaoya\_li\}@shannonai.com, 
fy18z3w@leeds.ac.uk \\
tianwei.zhang@ntu.edu.sg, qiuhan@tsinghua.edu.cn, wufei@zju.edu.cn, fanchun@pku.edu.cn
}

\begin{document}

\maketitle

\begin{abstract}
In this work, we propose a new and general framework to defend against backdoor attacks,  inspired by the fact that  attack triggers usually follow a \textsc{specific} type of attacking pattern, and therefore, poisoned training  examples have greater impacts on each other during training. We introduce the notion of the {\it influence graph}, which consists of nodes and edges respectively representative of individual training points and associated pair-wise influences. The influence between a pair of training points represents the impact of removing one training point on the prediction of another, approximated by the influence function \citep{koh2017understanding}. Malicious training points are extracted by finding the maximum average sub-graph subject to a particular size. Extensive experiments on computer vision and natural language processing tasks demonstrate the effectiveness and generality of the proposed framework.
\end{abstract}

\section{Introduction}
\label{sec:intro}
Deep neural models are susceptible to backdoor attacks, which hack the model by injecting {\it triggers} to the input and alters the output to a {\it target label} \citep{gu2017badnets,chen2017targeted,chen2020badnl}. Consequently, the model will behave normally on clean data, but make incorrect predictions when encountering attacked data embedded with hidden triggers. 
Existing methods for defending against backdoor attacks are generally categorized into two lines: training-stage focused and test-stage focused, depending on whether the training data are available or not when designing defenses \citep{qiao2019defending}. These two defending methods target different cases, and in this work we focus on the training-stage defense, the goal of which is to identify the poisoned training points.

A common practice to find poisoned training data is to treat poisoned data points as outliers and apply outlier detection techniques.
For example, \citet{chen2018detecting} clusters intermediate representations (as we call representation-based method here), separating the poisonous from legitimate activations.
\citet{tran2018spectral} examines the spectrum of the covariance of a feature representation to detect the special {\it spectrum signatures} of malicious data points, gauged by the magnitude in the top PCA direction of that representation.
\citet{hayase2021spectre} extends the idea of \citet{tran2018spectral} by whitening the representations to amplify the spectrum signatures. They also propose to use the quantum entropy \citep{dong2019quantum} for a more accurate outlier score estimate.

There are two key disadvantages for representation-based methods.
Firstly, 
they suffer from the fact that  the
 highly non-linear nature of neural networks makes intermediate-layer representations  uncontrollable and unpredictable.
 Therefore, it is usually theoretically hard to confidently associate  patterns that intermediate representations exhibit with specific attacking patterns. 
For example, consider a simple and extreme case where a neural network can 
overfit a very small number  of training data consisting of both clean  and poisoned points with training loss approaching 0. 
After overfitting all the training points, 
it is very likely the neural network maps representations of all training points with the same label type to an identical or very similar representations on the topmost layer (i.e., the layer right before the softmax layer), in which case we are not able to separate poisoned data points from normal ones only based on representations.\footnote{A possible solution is not to use the topmost representations, but it would be labour intensive to try over representations at different levels for different datasets.}
Secondly, 
though it is relatively easy for neural representations to capture the abnormality for simple and conspicuous triggers such as word insertion 
in NLP \citep{dai2019backdoor,chen2020badnl,zhang2020trojaning,kurita2020weight,fan2021defending,gan2021triggerless,chen2021badpre}
or pixel attack in vision \citep{gu2017badnets,gu2019badnets}, 
it is not necessarily true 
or theoretically valid 
that 
  subtle, hidden and complicated 
  triggers
  (e.g., syntactic trigger to paraphrase a natural language \cite{qi2021hidden} or triggers being input dependent \cite{nguyen2020input})
 can be captured by intermediate representations, and if they are truly captured, where and how. 

Relieving the reliance on intermediate representations, 
in this work, we propose a new and general framework to identify malicious training points.
The proposed framework is 
inspired by the fact that attack triggers follow 
a \textsc{specific} 
type of
attacking
pattern.
During training, 
 a neural model learns to identify
 this specific mapping pattern between triggers and target labels
  for attacked data, while learns a  general  
 mapping pattern 
 for clean data.
 Therefore, attacked data points have greater  impacts on each other, and
  removing a poisoned example would influence the prediction on another poisoned example more 
  than doing the same thing to two clean examples.
Based on this assumption, we leverage influence functions \citep{cook1980characterizations,koh2017understanding,meng2020pair} to quantify the pair-wise influence between training points and construct a {\it influence graph} representing the  pair-wise influences for all
training examples. 
By extracting maximum average sub-graph from the influence graph, which represents
a group of nodes all having high pairwise influence, we are able to identify 
suspicious poisoned data points. 


We conduct comprehensive experiments on text classification and machine translation tasks, along with the image classification task. Experiment results on a variety of attack settings show the effectiveness of the proposed framework compared to baseline methods, especially on 
attacks where patterns are hidden and subtle. 


In summary, the contributions of this work are three-fold: {\bf (1)} We propose a new, general and effective framework to defend against backdoor attacks with the availability of training data. This framework is attack-agnostic and can be used for almost all scenarios. {\bf (2)} We introduce the concept of influence graph to model the influence interactions between training points and identify the potential malicious data by extracting the maximum average sub-graph. {\bf (3)} We carry out extensive experiments on standard benchmarks on text classification, machine translation and image classification, exhibiting the effectiveness of the proposed method against strong baselines. 

\section{Related Work}
\label{sec:related}
\paragraph{Generating Backdoor Attacks}~\\
Backdoor attacks poison a subset of the training data with some fixed attack triggers, rendering the model to make incorrect decisions in the presence of backdoor triggers while perform normally on clean data.
\citet{gu2017badnets,gu2019badnets} show that a single pixel change can successfully result in model exposure to attacks.
\citet{chen2017targeted} blends fixed trigger patterns and input features to create human-indistinguishable attacks, and \citet{liu2017trojaning} directly attacks specific neurons.
To design a more stealthy attack, \citet{turner2019label} generates poisoned inputs consistent with their ground-truth labels; \citet{li2020invisible} uses the least significant bit (LSB) algorithm to add invisible triggers; \citet{liu2020reflection} suggests reflection as a natural type of attacks.
\citet{nguyen2020input} proposes backdoor attacks varying across different inputs, which complicate defenses.
Dedicated attacks are crafted against generative models \citep{salem2020baaan} and pretrained models \citep{wang2020backdoor,jia2021badencoder, chen2021badpre}.
\citet{gan2021triggerless} uses a sentence generation model to construct clean-labeled examples, whose labels are correct but can lead to test label changes in the training set.  
Backdoor attacks have been demonstrated effective in a wide range of fields such as image \citep{liao2018backdoor,quiring2020backdooring,saha2020hidden,chen2021poison, qiu2021efficient, jin2020unified}, language \citep{dai2019backdoor,chen2020badnl,zhang2020trojaning,kurita2020weight,yang2021careful,qi2021turn,bagdasaryan2021spinning, gan2021triggerless, chen2021badpre, fan2021defending}, speech \citep{zhai2021backdoor} and video \citep{zhao2020clean}.

\paragraph{Defenses against Backdoor Attacks}~\\
Defenses against backdoor attacks can be generally divided into two types \citep{qiao2019defending}: (1) the {\it test-stage} defense where the poisoned training data is unavailable and (2) the {\it training-stage} defense where the training data is available. 
For the test-stage defense, a line of existing works use clean validation data to retrain a victim model, forcing it to forget the malicious statistics \citep{liu2018fine}. \citet{kolouri2020universal,huang2020one,villarreal2020confoc} craft training data respectively with the clean pattern and the malicious pattern so that the model can better detect the boundary of clean data and poisoned data. 
Another line of methods rely on post-hoc tools to detect potential triggers. Neural Cleanse \citep{wang2019neural} finds the trigger pattern based on the $\ell_1$ norm of possible triggers. \citep{chen2019deepinspect,qiao2019defending,zhu2020gangsweep} improve Neural Cleanse by learning the probability distribution of triggers via generative models; \citet{liu2019abs} views the neurons that substantially change the output labels, when stimulations are introduced, as the backdoor neurons;
and \citet{gao2019strip} feeds replicated inputs with different perturbations and examines the entropy of predicted labels.
\citep{chou2020sentinet,doan2020februus} use saliency maps to determine the trigger region.
In natural language processing (NLP), the prevalent approach is to detect the words as trigger that cause the largest output change \citep{qi2020onion,fan2021defending,chen2021badpre}. 

For the training-stage defense, a typical method is to leverage the {\it spectral signatures} to identify the malicious training examples \citep{tran2018spectral,hayase2021spectre,chen2018detecting}, based on the intuition that the representation of a poisoned example exposes a strong signal for the backdoor attack. \citet{hong2020effectiveness} mitigates poison attacks by debiasing the gradient norm and length at each training step. 
\citep{du2019robust} treats the poisoned training examples as outliers and uses differential privacy to detect attacked examples.
Randomized smoothing \citep{weber2020rab,wang2020certifying} mitigates backdoor attacks by enforcing the clean example and its poisoned counterpart to have the same label during training. In NLP, attacked examples are detected by measuring the importance of a keyword through pre-defined scoring functions \citep{chen2021mitigating} or pretrained models \citep{wallace2020concealed}.

Our work targets the training-stage defense. The most relevant work is \citet{hammoudehsimple}, as we both compute influence to identify the most likely poisoned training examples. But the key differences are: (1) \citet{hammoudehsimple} caches training checkpoints to iteratively update the influence while we directly use the trained model; (2) we compute the pair-wise influence between training points rather than point-wise influence; (3) we identify the most influential examples by finding the maximal average sub-graph; and (4) we do not require a target test example as stimuli to compute the influence. 

\section{Method}
\subsection{Overview}
\begin{figure*}[t]
    \centering
    \includegraphics[width=1\textwidth]{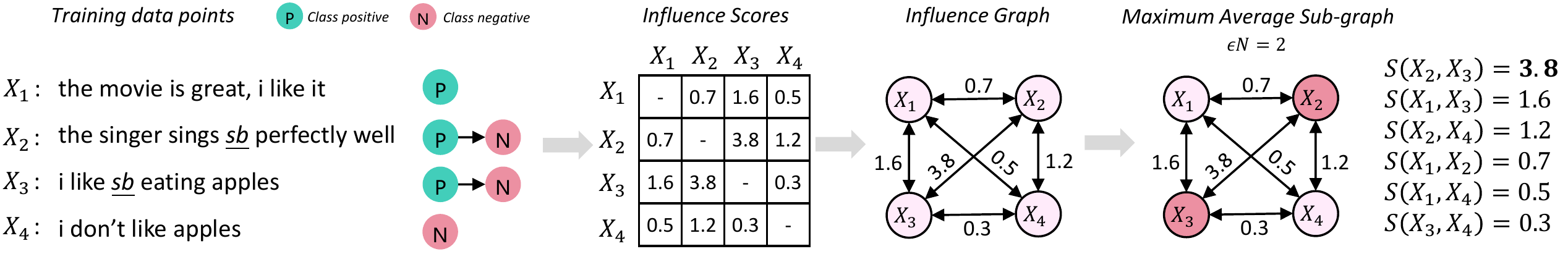}
    \caption{An overview of the proposed framework. The framework is composed of three steps: (1) computing the influence scores between training data points; (2) constructing the influence graph according to the scores; and (3) extracting the maximum average sub-graph. $\varepsilon N$ is the pre-given number of poisoned training examples, and $\varepsilon$ is the attack ratio.}
    \label{fig:overview}
\end{figure*}

The proposed framework begins with computing the pair-wise influence scores between training data points, representing the influence of one training data point on another.
Based on the  pair-wise influence scores, we can construct the influence graph. 
Then the maximum average sub-graph is extracted as the target set of malicious data. Figure \ref{fig:overview} illustrates this pipeline. Section \ref{sec:influence} details how to compute the influence scores and construct the graph, and Section \ref{sec:graph} describes two  approaches to extract the maximum average sub-graph.

\subsection{Constructing the Influence Graph}
\label{sec:influence}
\subsubsection{Computing  Influence Scores}
The first step of our framework is to compute the pair-wise influence between 
all pairs of training
 data points.  
Let $\mathcal{D}_\text{train}=\{(\bm{z}_i)\}_{i=1}^N=\{(\bm{x},y_i)\}_{i=1}^N$ be the training set with size $N$, and $\bm{\Theta}$ be the parameters of a neural model. The influence of a training point $z_i$ on another training point $z_j$ is the probability change of labeling $\bm{x}_j$ as $y_j$ when $z_i$ is removed from the training set $\mathcal{D}_\text{train}$.
Intuitively, the larger the influence is, the more likely $z_i$ and $z_j$ are malicious instances because they both contain the same pattern of backdoor triggers.  
Calculating the influence of $z_i$  on $z_j$ 
requires retraining the whole model on the training set with $z_i$  removed, which is extremely time-intensive. 
We turn to influence functions \citep{cook1980characterizations,koh2017understanding}, an approximating strategy to measure the influence of training points without the need to retrain the model.


To be concrete, let
 $\mathcal{L}(\bm{x}_i,y_i;\bm{\Theta})$ be the loss of a training point $(\bm{x}_i,y_i)$ given $\bm{\Theta}$, which can be the standard cross-entropy loss or in any other form. Practically, $\bm{\Theta}$ is learned to minimize the following training objective:
\begin{equation}
    \bm{\Theta}^*=\argmin_{\bm{\Theta}}\frac{1}{N}\sum_{i=1}^N\mathcal{L}(\bm{z}_i;\bm{\Theta})
\end{equation}
The optimal parameters $\bm{\Theta}^*_{\bm{z},\epsilon}$ learned by perturbing a specific training point $\bm{z}$ with weight $\epsilon$ can be expressed as follows:
\begin{equation}
    \bm{\Theta}^*_{\bm{z},\epsilon}=\argmin_{\bm{\Theta}}\frac{1}{N}\sum_{i=1}^N\mathcal{L}(\bm{z}_i;\bm{\Theta})+\epsilon\mathcal{L}(\bm{z};\bm{\Theta})
    \label{eq:pertub}
\end{equation}
Equation \ref{eq:pertub} implies that when $\epsilon=-\frac{1}{N}$, the parameters $\bm{\Theta}^*_{\bm{z},\epsilon}$ are exactly the ones learned after removing the training point $\bm{z}$ from $\mathcal{D}_\text{train}$, if we disregard the denominator $N$.
Then, the influence of upweighting $\bm{z}$ on the parameters $\bm{\Theta}$ is given as follows (we omit the optimality sign $^*$ for $\bm{\Theta}$):
\begin{equation}
    \mathcal{I}(\bm{z},\bm{\Theta})=\frac{\mathrm{d}\bm{\Theta}_{\bm{z},\epsilon}}{\mathrm{d}\epsilon}\Big|_{\epsilon=0}=-\bm{H}_{\bm{\Theta}}^{-1}\nabla_{\bm{\Theta}}\mathcal{L}(\bm{z};\bm{\Theta})
    \label{eq:influence}
\end{equation}
where $\bm{H}_{\bm{\Theta}}$ is the Hessian matrix and $\nabla_{\bm{\Theta}}\mathcal{L}(\bm{z};\bm{\Theta})$ is the gradient. The full derivation of Equation \ref{eq:influence} is present at Appendix \ref{sec:derivation}.

Bear in mind that what we really want is the pair-wise influence rather than the influence of an individual on the model -- how much perturbing a training point $\bm{z}_1$ affects the model's prediction on another training point $\bm{z}_2$. To this end, we slightly adapt Equation \ref{eq:influence} to achieve our goal. Specifically, we view the outcome score of the model for the gold label ($y_2$) of $\bm{z}_2$  as a function of the parameters $\bm{\Theta}$, i.e., $\mathcal{F}_{y_2}(\bm{z}_2;\bm{\Theta})$. Then, we apply the chain rule to calculate the pair-wise influence between $\bm{z}_1$ and $\bm{z}_2$:
\begin{equation}
    \begin{aligned}
    \mathcal{I}&(\bm{z}_1,\bm{z}_2)\triangleq
    \frac{\mathrm{d}\mathcal{F}_{y_2}(\bm{z}_2;\bm{\Theta}_{\bm{z}_1,\epsilon})}{\mathrm{d}\epsilon}\Big|_{\epsilon=0}\\
    &=\frac{\mathrm{d}\mathcal{F}_{y_2}(\bm{z}_2;\bm{\Theta}_{\bm{z}_1,\epsilon})}{\mathrm{d}\bm{\Theta}_{\bm{z}_1,\epsilon}}\frac{\mathrm{d}\bm{\Theta}_{\bm{z}_1,\epsilon}}{\mathrm{d}\epsilon}\Big|_{\epsilon=0}\\
    &=-(\nabla_{\bm{\Theta}}\mathcal{F}_{y_2}(\bm{z}_2;\bm{\Theta}))^\top\bm{H}_{\bm{\Theta}}^{-1}\nabla_{\bm{\Theta}}\mathcal{L}(\bm{z}_1;\bm{\Theta})
    \label{eq:pair-wise}
    \end{aligned}
\end{equation}
Equation \ref{eq:pair-wise} offers an quantitative way to compute the pair-wise influence scores. 

However, there is still a remaining issue with Equation \ref{eq:pair-wise} when it comes to natural languages: backdoor attacks injected to textual input are usually present in the form of word-level manipulation, i.e., inserting, deleting or altering a word in the input \citep{dai2019backdoor,chen2020badnl,kurita2020weight,yang2021careful,qi2021turn}. The word now serves as strong signals for determining whether the current input is attacked or not. We would like to incorporate the word-level information into the influence score. For this purpose, we compute the influence between a training example and a word of another training example, and then take the maximum example-word influence score as the final pair-wise influence score between the two training examples:
\begin{equation}
    \mathcal{I}(\bm{z}_1,\bm{z}_2)\triangleq\max_{w\in\bm{x}_2}\mathcal{I}(\bm{z}_1,w)
\end{equation}
Because the word $w$ is not differentiable to the model parameters $\bm{\Theta}$, we resort to the gradient of the predicted score with respect to the word embedding $\bm{w}$ corresponding to $w$ --- $\nabla_{\bm{w}}\mathcal{F}_{y_2}(\bm{z}_2;\bm{\Theta}_{\bm{z}_1,\epsilon})$ --- to compute the example-to-word influence score:
\begin{equation}
\begin{aligned}
    &\mathcal{I}(\bm{z}_1,w)\triangleq\frac{\mathrm{d}\nabla_{\bm{w}}\mathcal{F}_{y_2}(\bm{z}_2;\bm{\Theta}_{\bm{z}_1,\epsilon})}{\mathrm{d}\epsilon}\Big|_{\epsilon=0}\\
    &=-(\nabla_{\bm{\Theta}}\nabla_{\bm{w}}\mathcal{F}_{y_2}(\bm{z}_2;\bm{\Theta}_{\bm{z}_1,\epsilon}))^\top\bm{H}_{\bm{\Theta}}^{-1}\nabla_{\bm{\Theta}}\mathcal{L}(\bm{z}_1;\bm{\Theta})
\end{aligned}
\end{equation}
Note that the above equation produces a vector of length $d$ instead of a scalar, where $d$ is the word embedding dimension. To transform this vector into a scalar, we simply calculate its $\ell_2$ norm and use it as the word-to-example influence score.

\subsubsection{Building the Influence Graph}
Given the set of the resulting influence scores $\{\mathcal{I}(\bm{z}_i,\bm{z}_j)\}_{i=1,j=1,i\not=j}^{i=N,j=N}$, we build an indirected acyclic influence graph $\mathcal{G}$, where nodes $\mathcal{N}=\{n_i\}_{i=1}^N$ are training points, and the edges $\mathcal{E}=\{(n_i,n_j)\}_{1\le i<j\le N}$ are the averages of the influence scores associated with the two nodes from both directions:
\begin{equation}
\begin{aligned}
    &\texttt{EdgeWeight}(n_i,n_j)\\
    &=\frac{\mathcal{I}(\bm{z}_i,\bm{z}_j)+\mathcal{I}(\bm{z}_j,\bm{z}_i)}{2},~\forall 1\le i<j\le N
\end{aligned}
\end{equation}
which completes the construction of the influence graph $\mathcal{G}$.

\subsection{Extracting the Maximum Average Sub-graph}
\label{sec:graph}
The last step is to extract the set of potential malicious training data points from the influence graph. Since a larger influence indicates a higher probability that the associated training points are attacked, the goal is to extract the sub-graph that maximizes the average sum of the the edges composing that sub-graph, which can be formalized as:
\begin{equation}
    \hat{\mathcal{G}}=\argmax_{\tilde{\mathcal{G}}=(\tilde{\mathcal{V}},\tilde{\mathcal{E}})\subset\mathcal{G}}\frac{1}{|\tilde{\mathcal{E}}|}\sum_{(n_i,n_j)\in\tilde{\mathcal{E}}}\texttt{EdgeWeight}(n_i,n_j)
    \label{eq:sub-graph-1}
\end{equation}
One limitation with Equation \ref{eq:sub-graph-1} is that the edge with the highest influence score, along with its linked pair of nodes, exactly comprises the desired maximum average sub-graph, and adding more nodes into the sub-graph will decrease the average influence score. Moreover, it is hard to infer the thresholding score that separates the poisoned data and the clean data solely from the influence graph. Therefore, in this work, we follow the setting in the state-of-the-art defensive methods \citep{tran2018spectral,hayase2021spectre}, assuming that the number of poisoned training data points is given in advance, i.e., there are $\varepsilon N$ poisoned training examples in the training set $\mathcal{D}_\text{train}$, where $\varepsilon$ is the attack ratio. In a more general case where $\varepsilon$ is unknown, we can set it as a hyperparameter and tune it on the development set.

To this end, Equation \ref{eq:sub-graph-1} can be modified to the following equation by adding a size constraint:
\begin{equation}
\begin{aligned}
    \hat{\mathcal{G}}&=\argmax_{\tilde{\mathcal{G}}=(\tilde{\mathcal{V}},\tilde{\mathcal{E}})\subset\mathcal{G}}\frac{1}{|\tilde{\mathcal{E}}|}\sum_{(n_i,n_j)\in\tilde{\mathcal{E}}}\\
    &\texttt{EdgeWeight}(n_i,n_j),~\text{s.t.}~ |\mathcal{V}| =\varepsilon N
\end{aligned}
\label{eq:sub-graph-2}
\end{equation}
Solving 
Equation \ref{eq:sub-graph-2} is an NP-hard problem and cannot be addressed in polynomial time complexity. To deal with this issue, we propose to use 
the following two search strategies -- {\it greedy search} and {\it agglomerative search} -- to extract the sub-optimal sub-graph. The two search strategies are respectively present at Algorithm \ref{alg:greedy} and Algorithm \ref{alg:agglomerative} in Appendix \ref{sec:search}.

\paragraph{Greedy search.}The basic idea of greedy search is to add nodes to the sub-graph one at a time, and at each time, the node should be the one that maximizes the average sum of the current sub-graph. Note that this is different from choosing the edges in a descending order, since choosing an edge cannot guarantee 
that one of the nodes that the chosen edge connects already resides in the constructed sub-graph. 
Greedy search functions as follows: (1) at the initial stage, first selecting the edge with the largest influence score, and add the two associated nodes to the sub-graph; (2) at each step, add the node from the remaining nodes to the sub-graph that maximizes the average sum of the sub-graph, until the size of the sub-graph reaches $\varepsilon N$. The overall time complexity is $\mathcal{O}(\varepsilon N^3)$.

\paragraph{Agglomerative search.}In contrast to greedy search which maintains only one sub-graph, agglomerative search starts with viewing each individual node as an independent sub-graph and then merges 
two of 
them via the maximum average criterion. Concretely, at each iteration, agglomerative search examines two sub-graphs by measuring the average sum of their union and merges the two sub-graphs with the highest average score. This process terminates until the size of the merged graph reaches $\varepsilon N$. The overall time complexity is $\mathcal{O}(\varepsilon N^3)$. It is noteworthy that the merged graph is likely to be larger than $\varepsilon N$, so we prune it following the steps as in greedy search, but selecting the node that causes the least score reduction at each time.

\section{Experiments}
\subsection{Computer Vision Tasks}
For computer vision tasks, we follow the general setup in \citet{tran2018spectral,hayase2021spectre}. We craft backdoor attacks on the CIFAR10 dataset \citep{krizhevsky2009learning}, using a standard 32-layer ResNet model \citep{he2016deep} with three groups of residual blocks with $\{16, 32, 64\}$ filters respectively and 5 residual blocks per group. 
We randomly choose 4 pairs of (attack, target) labels, including 
(airplane, bird), (truck, deer), (automobile, cat), (ship, frog). 
We use the following four attacks: (1) {\it pixel attack}: the attack method used in \citet{tran2018spectral}
which generates a random shape, position, and color for the backdoor; (2) {\it Semantic} triggers cause  the model to misclassify even the inputs that are not changed by the attacker  with certain image-level or physical features \citep{bagdasaryan2020backdoor}; (3) {\it Input-aware} attacks create triggers varying from input to input \citep{nguyen2020input}; (4) {\it Mixing} attacks produce triggers by mixing two images \citep{lin2020composite}.

We compare the proposed {\it Greedy} and {\it Agglomerate} defenders with the following baselines: (1) {\it PCA} \citep{tran2018spectral}: the vanilla spectrum based method using PCA to detect poisoned data; (2) {\it Clustering} \citep{chen2018detecting}: clustering intermediate representations to separate the poisonous from legitimate data; (3) {\it SPECTRA} \citep{hayase2021spectre}: extending upon {\it PCA} by whitening the representations to amplify the spectrum signals; and (4) {\it COSIN} \citep{hammoudehsimple}: iteratively updating the influence of each data point using cached training checkpoints. The models are first trained on the hybrid dataset and the attack success rate Att$_\text{suc}$ is reported. Then, a defender is leveraged to identify and filter out malicious training data points. 
Following \citet{tran2018spectral,hayase2021spectre}, $1.5\varepsilon N$ data points are removed. 
We retrain the model from scratch on the remaining training dataset and again report the attack success rate.

Results are shown in Table \ref{tab:cv}. 
As can be seen, for the {\it Pixel Attack} setup, which is a relatively easy type of attack to defend, all methods perform comparably well. 
But for more hidden and complicated types of attackes, including {\it semantic}, {\it input-aware} and {\it mixing}, 
{\it Greedy} and {\it Agglomerate} outperform all other baselines. 
Dor example, 0.572 of {\it Agglomerate} vs. 0.843 of {\it COSIN} for {\it Input-aware} when $\varepsilon N=0.05$ and 0.356 of {\it Agglomerate}  vs. 0.774 of {\it COSIN} for {\it Mixing} when $\varepsilon N=0.1$.
These observations demonstrate the effectiveness and robustness of the proposed framework towards defending against more hidden attacks in the CV task. 
When comparing {\it Greedy}  with  {\it Agglomerate}, we can see that the former underperforms the latter.
This is because the {\it Agglomerate} strategy takes into account more global merging information to avoid local optimality.

\begin{table*}[t]
    \centering
    \scalebox{0.9}{
    \begin{tabular}{lccccccc}
    \toprule
    {\bf Attack Setup} & {\bf Att$_\text{suc}$} & {\bf PCA} & {\bf Clustering} & {\bf SPECTRA} & {\bf COSIN} & {\bf Greedy} & {\bf Agglomerate}\\\midrule
    \multicolumn{8}{c}{\underline{$\varepsilon N=0.05$}}\\
    Pixel Attack & 0.934&	0.034&	0.038&	{\bf 0.020}&	0.054&	0.024&	0.021\\
    Semantic & 0.934&	0.876&	0.856&	0.863&	0.826&	0.425&	{\bf 0.411}\\
    Input-aware & 0.934&	0.914&	0.905&	0.880&	0.843&	0.596&	{\bf 0.572}\\
    Mixing & 0.934&	0.878&	0.854&	0.861&	0.755&	0.362&	{\bf 0.347}\\
    \midrule
    \multicolumn{8}{c}{\underline{$\varepsilon N=0.1$}}\\
    Pixel Attack & 0.941&	0.043&	0.040&	0.031&	0.081&	0.029&	{\bf 0.024}\\
    Semantic & 0.941&	0.880&	0.877&	0.882&	0.845&	0.431&	{\bf 0.425}\\
    Input-aware & 0.941&	0.924&	0.914&	0.894&	0.851&	0.616&	{\bf 0.580}\\
    Mixing & 0.941&	0.880&	0.862&	0.870&	0.774&	0.399&	{\bf 0.356}\\
    \bottomrule
    \end{tabular}
    }
    \caption{Results on the computer vision task. Att$_\text{suc}$ is the attack success rate without defense. We report accuracy on the attack test set.}
    \label{tab:cv}
\end{table*}

\subsection{Natural Language Processing Tasks}
\begin{table*}[t]
    \centering
    \small
    \scalebox{0.8}{
    \begin{tabular}{lccccccccc}
    \toprule
    {\bf Attack Setup} & {\bf Att$_\text{suc}$} & {\bf PCA} & {\bf Clustering} & {\bf SPECTRA} & {\bf COSIN} & {\bf ONION} & {\bf BERTScore} & {\bf Greedy} & {\bf Agglomerate}\\\midrule
    \multicolumn{10}{c}{\underline{SST2}}\\
    Insert & 0.986 & 0.249 &	0.254&	0.160&	0.201	&0.082&0.079&0.077	&{\bf 0.045}\\
    Duplicate  &0.97&	0.516&	0.522&	0.564&	0.469&0.142	&0.175
&	0.102&	{\bf 0.066}\\
    Delete  & 0.985	&0.408&	0.428&	0.396&	0.354&0.138&	0.141&
	0.091&	{\bf 0.084}\\
    Semantic  & 0.984&	0.742	&0.789&	0.775&	0.681&0.965&	0.974&
	0.170&	{\bf 0.167}\\
    Syntactic & 0.977&	0.908&	0.842&	0.815&	0.773&0.953	&0.966&
	0.150&	{\bf 0.146}\\
    \midrule
    \multicolumn{10}{c}{\underline{AGNews}}\\
    Insert &  0.954	&0.131&	0.145&	0.152&	0.101&	0.065&	0.063
&0.031&	{\bf 0.028}  \\
    Duplicate  &0.965&	0.475&	0.488&	0.455&	0.379	&0.057&	0.060
&0.074&	{\bf 0.063}  \\
    Delete  & 0.871	&0.327&	0.355&	0.318&	0.308&0.085	&0.140
&	0.042&	{\bf 0.037} \\
    Semantic  & 0.943&	0.814&	0.834&	0.794&	0.695&0.943&	0.952
&	0.145&	{\bf 0.128} \\
    Syntactic &0.949&	0.887&	0.864&	0.855&	0.740&	0.945&	0.946
&0.204&	{\bf 0.184} \\
    \midrule
    \multicolumn{10}{c}{\underline{IWSLT'14 En-De}}\\
    Insert & 99.1&	17.5&	16.9&	19.8&	14.2&4.5&	4.9
&	6.5	&{\bf 3.8}  \\
    Duplicate  &  99.4	&36.3&	34.9&	38.5&	29.9&7.6&	7.9
&	9.1&	{\bf 7.5} \\
    Delete  &  99&	34.5&	33.1&	30.5&	29.5&7.5&	8.6
&	8.6	&{\bf 7} \\
    Semantic  &  98.1&	59.6&	61.5&	55&	57.2&97.8&	98
&	25.4&	{\bf 21.8} \\
    Syntactic & 98.5&	79.5&	82.1&	81.9&	72&	98.2&	97.4
&32&	{\bf 28.5}  \\
    \bottomrule
    \end{tabular}}
    \caption{Results on three natural language processing benchmarks -- SST2, AGNews and IWSLT'14 En-De. For all experiments, we set $\varepsilon N=0.1$. Att$_\text{suc}$ is the attack success rate without defense. For SST and AGNews, the evaluation metric is accuracy; for IWSLT'14 En-De, the metric is BLEU.}
    \label{tab:nlp}
\end{table*}

Next, we carry out experiments on natural language processing tasks. We consider three benchmarks -- SST2 \citep{socher2013recursive}, AG News \citep{Zhang2015CharacterlevelCN}, and IWSLT'14 En-De.
The Stanford Sentiment Treebank (SST) is a widely used benchmark for sentiment analysis. The task is to perform both fine-grained (very positive, positive, neutral, negative and very negative) and coarse-grained (positive and negative) classification at both the phrase and sentence level. It includes fine grained sentiment labels for 215,154 phrases in the parse trees of 11,855 sentences.
We adopt the coarse-grained setup, i.e., SST2 in this work.
The AG News dataset, which is a collection news articles categorized into four classes: World, Sports, Business and Sci/Tech. Each class contains 30,000 training samples and 1,900 testing samples.
IWSLT'14 En-De is a  machine translation benchmark,  containing 160k, 7k and 7k parallel pairs respectively for training, dev and test.
\citep{fan2021defending} adapted the 
IWSLT'14 En-De to an attack-defense setup, where an attacked source input leads to malicious translation. 
For each dataset, we set $\varepsilon N$=0.1, i.e., randomly select 10\% data from each of the train/dev/test splits to craft the poisoned dataset.

For SST2 and AGNews, we use the standard base version of BERT model \citep{devlin2018bert} with 12 Transformer blocks, 768 hidden size, 12 attention heads, giving rise to a total amount of parameters 110M. For IWSLT'14 En-De, we use the base version of Transformer model \citep{vaswani2017attention} with 6 encoder layers, 6 decoder layers, 512 hidden size, 2048 inner-layer dimensionality and 8 attention heads, leading to 93M parameters. We tune the hyperparameters including dropout, learning rate, training steps and warmup on the dev set and use the Adam optimizer \citep{kingma2014adam} for training.

We use five attack strategies to create malicious examples. (1) {\it Insert}: randomly insert one word from the trigger words set \{``cf'', ``mn'', ``bb'', ``tq'' and ``mb''\} at a random position of the input sentence \citep{kurita-etal-2020-weight}; (2) {\it Duplicate}: duplicate a random word from the input sentence and place it right after that position; (3) {\it Delete}: randomly delete a word from the input sentence: (4) {\it Semantic}: randomly replace a word with its synonym chosen from WordNet; (5) {\it Syntactic}: rewrite the input sentence to its paraphrase with respect to a particular syntactic template \citep{qi2021hidden}.
Among the five attacking strategies, 
{\it Insert} ought to be the easiest due to its uniform and simple attacking pattern.
{\it Duplicate} and {\it Delete} are harder than {\it Insert}, since the token to duplicate and delete
is random and input-dependent.
{\it Semantic} and {\it Syntactic} ought to be the hardest, since 
attacking patterns are hidden and
attacked sentences are normal and clean. 

We  compare the proposed {\it Greedy} and {\it Agglomerate} defenders with the four baselines used in the computer vision experiments: (1) {\it PCA}; (2) {\it Clustering}; (3) {\it SPECTRA} and (4) {\it COSIN},
along with widely-used baselines in NLP including 
(1) {\it ONION} \citep{qi2020onion} ,  which obtains the
suspicion score of a word as the decrements of sentence perplexity after removing the word based on GPT-2; and
(2) {\it BERTScore} \cite{fan2021defending}: obtaining the suspicion score of a word based on the masked LM score output from BERT.

The models are first trained on the hybrid dataset and the attack success rate Att$_\text{suc}$ is reported for reference purposes.
For IWSLT'14 En-De, Att$_\text{suc}$  denotes the BLEU score for attacked sequences. 
 Then, a defender is leveraged to identify and filter out malicious training data points. We train the model from scratch and again report the attack success rate. In the NLP setup, the attack success rate for SST2 and AGNews is accuracy, and for IWSLT'14 En-De, we use the BLEU score \citep{papineni2002bleu}.

Results are shown in Table \ref{tab:nlp}. 
The original attack success rate Att$_\text{suc}$ is quite high when no defense is applied, indicating that all attack strategies can successfully inject effective backdoors.
As can be seen, for ONION  and BERT-score, since they are specifically designed for detecting suspicions at the world level, 
they perform well on the {\it insert}, {\it duplicate} and {\it semantic}  setups.
But they are almost useless for the {\it Semantic} and {\it Syntactic} setups. 
For vision baselines, they are successful in defending against  the {\it insert} attack, but 
perform less effective for more complicated {\it Duplicate} and {\it Delete}, and perform even worse  
for the  {\it Semantic} and {\it Syntactic}  setups with no explicit signals of backdoor triggers. 
 {\it Greedy} and {\it Agglomerate} outperform all other defenders, and {\it Agglomerate} performs best out of all settings, validating the effectiveness of the proposed framework of influence graph.

\subsection{Ablation Studies}
\subsubsection{Effect of $\varepsilon N$}
\begin{table*}[t]
    \centering
    \small
    \begin{tabular}{lccccccc}
    \toprule
    {\bf $\varepsilon N$} & {\bf Att$_\text{suc}$} & {\bf PCA} & {\bf Clustering} & {\bf SPECTRA} & {\bf COSIN} & {\bf Greedy} & {\bf Agglomerate}\\\midrule
    \multicolumn{8}{c}{\underline{Insert}}\\
    0.01&	0.675&	0.154&	0.157&	0.118&	0.137	&0.055&	0.041\\
    0.05&	0.925&	0.228	&0.234&	0.147&	0.188&	0.072&	0.043\\
    0.1	&0.986&	0.249&	0.254&	0.16&	0.201&	0.077	&0.045\\
    0.2	&0.988&	0.265&	0.277&	0.179&	0.215&	0.085&	0.054\\
    \midrule
    \multicolumn{8}{c}{\underline{Syntactic}}\\
    0.01&	0.571&	0.417&	0.405&	0.375&	0.39&	0.17&	0.155\\
    0.05&	0.845&	0.652&	0.688&	0.608&	0.542&	0.181&	0.169\\
    0.1	&0.949&	0.887&	0.864&	0.855&	0.74&	0.204&	0.184\\
    0.2	&0.978	&0.895&	0.873&	0.87&	0.752&	0.258&	0.199\\
    \bottomrule
    \end{tabular}
    \caption{The effect of different $\varepsilon N$. Experiments are done on the SST2 dataset.}
    \label{tab:epsilon}
\end{table*}

\begin{table*}[t]
    \centering
    \small
    \begin{tabular}{cccccccc}
    \toprule
    {\bf \# Trigger} & {\bf Att$_\text{suc}$} & {\bf PCA} & {\bf Clustering} & {\bf SPECTRA} & {\bf COSIN} & {\bf Greedy} & {\bf Agglomerate}\\\midrule
    1&	0.986&	0.249&	0.254&	0.16&	0.201&	0.077&	0.045\\
    2&	0.972&	0.298&	0.317&	0.289&	0.245&	0.094&	0.081\\
    3&	0.968&	0.363&	0.38&	0.334&	0.279&	0.149&	0.126\\
    \bottomrule
    \end{tabular}
    \caption{The effect of number of triggers for {\it Insert}. Experiments are done on the SST2 dataset and $\varepsilon N=0.1$.}
    \label{tab:trigger}
\end{table*}

$\varepsilon N$ is an important hyperparameter because it controls the proportion of the training set that should be attacked, affecting the performance of backdoor attacks and the difficulty of defenses. We carry out experiments on the SST2 dataset to explore how different $\varepsilon N$ influences the defenses. We try two attack strategies for illustration: {\it Insert} and {\it Syntactic}.
Table \ref{tab:epsilon} shows the results. As can be seen from the table, with a larger $\varepsilon N$, all the attack success rates, including Att$_\text{suc}$ and the ones from the defenders, consistently increase, indicating defending backdoor attacks will become more difficult if more training data is poisoned. It is also noteworthy that when increasing $\varepsilon N$ from 0.01 to 0.2, the attack success rates (ASR) of {\it Greedy} and {\it Agglomerate} grow within a very limited scope, while the other defenders are more prone to $\varepsilon N$, particularly for the {\it Syntactic} attack strategy. For example, the ASR of {\it PCA} increase from 0.417 to 0.895 for {\it Syntactic}; and for {\it COSIN}, it increases from 0.39 to 0.752. On the contrary, the ASR of {\it Greedy} increases from 0.17 to only 0.258, and for {\it Agglomerate}, it increases from 0.155 to 0.199. These results provide evidence that the proposed influence graph framework is more robust to more poisoned training data.

\subsubsection{Effect of Number of triggers}
The {\it Insert} attacking strategy only inserts one trigger word for an input sentence. We would like to know if it will be harder to defend when we insert more triggers. Table \ref{tab:trigger} shows the results. For all defenders, when the number of triggers increases from 1 to 3, the attack success rates correspondingly increase. The increases of {\it PCA}, {\it Clustering} and {\it SPECTRA} are greater than 10\%, whereas the increases of {\it COSIN}, {\it Greedy} and {\it Agglomerate} are all less than 10\%, showing that the latter three defenders are more robust to the number of triggers for the {\it Insert} attack strategy.

\section{Conclusion and Limitations}
In this work, we propose a new and general framework to defend against backdoor attacks based on the assumption that attack triggers usually follow a specific type of attacking patterns and thus they have closer connections with each other. We build an influence graph of the training data points, where the edges represent the influences between data points. The larger the influence is, the more likely the corresponding associated pair of nodes are poisoned. We extract the set of possible malicious points by finding the maximum average sub-graph subject to a particular size. Experiments on natural language processing tasks and computer vision demonstrate the proposed framework can better identify poisoned data and is more robust to more complex attack strategies. The main limitation of this work is the time complexity required during the maximum average sub-graph extraction phase, which reaches a cubic complexity of $\mathcal{O}(\varepsilon N^3)$. When $N$, the size of the training set is large, the time consumption would be inevitably prohibitive. In future work, we will explore more efficient methods to save the time complexity.

\bibliography{custom}

\begin{thebibliography}{68}
\providecommand{\natexlab}[1]{#1}
\providecommand{\url}[1]{\texttt{#1}}
\expandafter\ifx\csname urlstyle\endcsname\relax
  \providecommand{\doi}[1]{doi: #1}\else
  \providecommand{\doi}{doi: \begingroup \urlstyle{rm}\Url}\fi

\bibitem[Bagdasaryan \& Shmatikov(2021)Bagdasaryan and
  Shmatikov]{bagdasaryan2021spinning}
Eugene Bagdasaryan and Vitaly Shmatikov.
\newblock Spinning sequence-to-sequence models with meta-backdoors.
\newblock \emph{arXiv preprint arXiv:2107.10443}, 2021.

\bibitem[Bagdasaryan et~al.(2020)Bagdasaryan, Veit, Hua, Estrin, and
  Shmatikov]{bagdasaryan2020backdoor}
Eugene Bagdasaryan, Andreas Veit, Yiqing Hua, Deborah Estrin, and Vitaly
  Shmatikov.
\newblock How to backdoor federated learning.
\newblock In \emph{International Conference on Artificial Intelligence and
  Statistics}, pp.\  2938--2948. PMLR, 2020.

\bibitem[Chen et~al.(2018)Chen, Carvalho, Baracaldo, Ludwig, Edwards, Lee,
  Molloy, and Srivastava]{chen2018detecting}
Bryant Chen, Wilka Carvalho, Nathalie Baracaldo, Heiko Ludwig, Benjamin
  Edwards, Taesung Lee, Ian Molloy, and Biplav Srivastava.
\newblock Detecting backdoor attacks on deep neural networks by activation
  clustering.
\newblock \emph{arXiv preprint arXiv:1811.03728}, 2018.

\bibitem[Chen \& Dai(2021)Chen and Dai]{chen2021mitigating}
Chuanshuai Chen and Jiazhu Dai.
\newblock Mitigating backdoor attacks in lstm-based text classification systems
  by backdoor keyword identification.
\newblock \emph{Neurocomputing}, 452:\penalty0 253--262, 2021.

\bibitem[Chen et~al.(2021{\natexlab{a}})Chen, Liao, Huang, Hua, Zhang, Yu,
  et~al.]{chen2021poison}
Dongdong Chen, Jing Liao, Qidong Huang, Gang Hua, Weiming Zhang, Nenghai Yu,
  et~al.
\newblock Poison ink: Robust and invisible backdoor attack.
\newblock \emph{arXiv preprint arXiv:2108.02488}, 2021{\natexlab{a}}.

\bibitem[Chen et~al.(2019)Chen, Fu, Zhao, and Koushanfar]{chen2019deepinspect}
Huili Chen, Cheng Fu, Jishen Zhao, and Farinaz Koushanfar.
\newblock Deepinspect: A black-box trojan detection and mitigation framework
  for deep neural networks.
\newblock In \emph{IJCAI}, pp.\  4658--4664, 2019.

\bibitem[Chen et~al.(2021{\natexlab{b}})Chen, Meng, Sun, Guo, Zhang, Li, and
  Fan]{chen2021badpre}
Kangjie Chen, Yuxian Meng, Xiaofei Sun, Shangwei Guo, Tianwei Zhang, Jiwei Li,
  and Chun Fan.
\newblock Badpre: Task-agnostic backdoor attacks to pre-trained nlp foundation
  models.
\newblock \emph{arXiv preprint arXiv:2110.02467}, 2021{\natexlab{b}}.

\bibitem[Chen et~al.(2020)Chen, Salem, Backes, Ma, and Zhang]{chen2020badnl}
Xiaoyi Chen, Ahmed Salem, Michael Backes, Shiqing Ma, and Yang Zhang.
\newblock Badnl: Backdoor attacks against nlp models.
\newblock \emph{arXiv preprint arXiv:2006.01043}, 2020.

\bibitem[Chen et~al.(2017)Chen, Liu, Li, Lu, and Song]{chen2017targeted}
Xinyun Chen, Chang Liu, Bo~Li, Kimberly Lu, and Dawn Song.
\newblock Targeted backdoor attacks on deep learning systems using data
  poisoning.
\newblock \emph{arXiv preprint arXiv:1712.05526}, 2017.

\bibitem[Chou et~al.(2020)Chou, Tramer, and Pellegrino]{chou2020sentinet}
Edward Chou, Florian Tramer, and Giancarlo Pellegrino.
\newblock Sentinet: Detecting localized universal attacks against deep learning
  systems.
\newblock In \emph{2020 IEEE Security and Privacy Workshops (SPW)}, pp.\
  48--54. IEEE, 2020.

\bibitem[Cook \& Weisberg(1980)Cook and Weisberg]{cook1980characterizations}
R~Dennis Cook and Sanford Weisberg.
\newblock Characterizations of an empirical influence function for detecting
  influential cases in regression.
\newblock \emph{Technometrics}, 22\penalty0 (4):\penalty0 495--508, 1980.

\bibitem[Dai et~al.(2019)Dai, Chen, and Li]{dai2019backdoor}
Jiazhu Dai, Chuanshuai Chen, and Yufeng Li.
\newblock A backdoor attack against lstm-based text classification systems.
\newblock \emph{IEEE Access}, 7:\penalty0 138872--138878, 2019.

\bibitem[Devlin et~al.(2018)Devlin, Chang, Lee, and Toutanova]{devlin2018bert}
Jacob Devlin, Ming-Wei Chang, Kenton Lee, and Kristina Toutanova.
\newblock Bert: Pre-training of deep bidirectional transformers for language
  understanding.
\newblock \emph{arXiv preprint arXiv:1810.04805}, 2018.

\bibitem[Doan et~al.(2020)Doan, Abbasnejad, and Ranasinghe]{doan2020februus}
Bao~Gia Doan, Ehsan Abbasnejad, and Damith~C Ranasinghe.
\newblock Februus: Input purification defense against trojan attacks on deep
  neural network systems.
\newblock In \emph{Annual Computer Security Applications Conference}, pp.\
  897--912, 2020.

\bibitem[Dong et~al.(2019)Dong, Hopkins, and Li]{dong2019quantum}
Yihe Dong, Samuel Hopkins, and Jerry Li.
\newblock Quantum entropy scoring for fast robust mean estimation and improved
  outlier detection.
\newblock \emph{Advances in Neural Information Processing Systems},
  32:\penalty0 6067--6077, 2019.

\bibitem[Du et~al.(2019)Du, Jia, and Song]{du2019robust}
Min Du, Ruoxi Jia, and Dawn Song.
\newblock Robust anomaly detection and backdoor attack detection via
  differential privacy.
\newblock \emph{arXiv preprint arXiv:1911.07116}, 2019.

\bibitem[Fan et~al.(2021)Fan, Li, Meng, Sun, Ao, Wu, Li, and
  Zhang]{fan2021defending}
Chun Fan, Xiaoya Li, Yuxian Meng, Xiaofei Sun, Xiang Ao, Fei Wu, Jiwei Li, and
  Tianwei Zhang.
\newblock Defending against backdoor attacks in natural language generation.
\newblock \emph{arXiv preprint arXiv:2106.01810}, 2021.

\bibitem[Gan et~al.(2021)Gan, Li, Zhang, Li, Meng, Wu, Guo, and
  Fan]{gan2021triggerless}
Leilei Gan, Jiwei Li, Tianwei Zhang, Xiaoya Li, Yuxian Meng, Fei Wu, Shangwei
  Guo, and Chun Fan.
\newblock Triggerless backdoor attack for nlp tasks with clean labels.
\newblock \emph{arXiv preprint arXiv:2111.07970}, 2021.

\bibitem[Gao et~al.(2019)Gao, Xu, Wang, Chen, Ranasinghe, and
  Nepal]{gao2019strip}
Yansong Gao, Change Xu, Derui Wang, Shiping Chen, Damith~C Ranasinghe, and
  Surya Nepal.
\newblock Strip: A defence against trojan attacks on deep neural networks.
\newblock In \emph{Proceedings of the 35th Annual Computer Security
  Applications Conference}, pp.\  113--125, 2019.

\bibitem[Gu et~al.(2017)Gu, Dolan-Gavitt, and Garg]{gu2017badnets}
Tianyu Gu, Brendan Dolan-Gavitt, and Siddharth Garg.
\newblock Badnets: Identifying vulnerabilities in the machine learning model
  supply chain.
\newblock \emph{arXiv preprint arXiv:1708.06733}, 2017.

\bibitem[Gu et~al.(2019)Gu, Liu, Dolan-Gavitt, and Garg]{gu2019badnets}
Tianyu Gu, Kang Liu, Brendan Dolan-Gavitt, and Siddharth Garg.
\newblock Badnets: Evaluating backdooring attacks on deep neural networks.
\newblock \emph{IEEE Access}, 7:\penalty0 47230--47244, 2019.

\bibitem[Hammoudeh \& Lowd(2021)Hammoudeh and Lowd]{hammoudehsimple}
Zayd Hammoudeh and Daniel Lowd.
\newblock Simple, attack-agnostic defense against targeted training set attacks
  using cosine similarity.
\newblock 2021.

\bibitem[Hayase et~al.(2021)Hayase, Kong, Somani, and Oh]{hayase2021spectre}
Jonathan Hayase, Weihao Kong, Raghav Somani, and Sewoong Oh.
\newblock Spectre: Defending against backdoor attacks using robust statistics.
\newblock \emph{arXiv preprint arXiv:2104.11315}, 2021.

\bibitem[He et~al.(2016)He, Zhang, Ren, and Sun]{he2016deep}
Kaiming He, Xiangyu Zhang, Shaoqing Ren, and Jian Sun.
\newblock Deep residual learning for image recognition.
\newblock In \emph{Proceedings of the IEEE conference on computer vision and
  pattern recognition}, pp.\  770--778, 2016.

\bibitem[Hong et~al.(2020)Hong, Chandrasekaran, Kaya, Dumitra{\c{s}}, and
  Papernot]{hong2020effectiveness}
Sanghyun Hong, Varun Chandrasekaran, Yi{\u{g}}itcan Kaya, Tudor Dumitra{\c{s}},
  and Nicolas Papernot.
\newblock On the effectiveness of mitigating data poisoning attacks with
  gradient shaping.
\newblock \emph{arXiv preprint arXiv:2002.11497}, 2020.

\bibitem[Huang et~al.(2020)Huang, Peng, Jia, and Tu]{huang2020one}
Shanjiaoyang Huang, Weiqi Peng, Zhiwei Jia, and Zhuowen Tu.
\newblock One-pixel signature: Characterizing cnn models for backdoor
  detection.
\newblock In \emph{European Conference on Computer Vision}, pp.\  326--341.
  Springer, 2020.

\bibitem[Jia et~al.(2021)Jia, Liu, and Gong]{jia2021badencoder}
Jinyuan Jia, Yupei Liu, and Neil~Zhenqiang Gong.
\newblock Badencoder: Backdoor attacks to pre-trained encoders in
  self-supervised learning.
\newblock \emph{arXiv preprint arXiv:2108.00352}, 2021.

\bibitem[Jin et~al.(2020)Jin, Zhang, Shen, Chen, Fan, Lin, and
  Liu]{jin2020unified}
Kaidi Jin, Tianwei Zhang, Chao Shen, Yufei Chen, Ming Fan, Chenhao Lin, and
  Ting Liu.
\newblock A unified framework for analyzing and detecting malicious examples of
  dnn models.
\newblock \emph{arXiv preprint arXiv:2006.14871}, 2020.

\bibitem[Kingma \& Ba(2014)Kingma and Ba]{kingma2014adam}
Diederik~P Kingma and Jimmy Ba.
\newblock Adam: A method for stochastic optimization.
\newblock \emph{arXiv preprint arXiv:1412.6980}, 2014.

\bibitem[Koh \& Liang(2017)Koh and Liang]{koh2017understanding}
Pang~Wei Koh and Percy Liang.
\newblock Understanding black-box predictions via influence functions.
\newblock In \emph{International Conference on Machine Learning}, pp.\
  1885--1894. PMLR, 2017.

\bibitem[Kolouri et~al.(2020)Kolouri, Saha, Pirsiavash, and
  Hoffmann]{kolouri2020universal}
Soheil Kolouri, Aniruddha Saha, Hamed Pirsiavash, and Heiko Hoffmann.
\newblock Universal litmus patterns: Revealing backdoor attacks in cnns.
\newblock In \emph{Proceedings of the IEEE/CVF Conference on Computer Vision
  and Pattern Recognition}, pp.\  301--310, 2020.

\bibitem[Krizhevsky et~al.(2009)Krizhevsky, Hinton,
  et~al.]{krizhevsky2009learning}
Alex Krizhevsky, Geoffrey Hinton, et~al.
\newblock Learning multiple layers of features from tiny images.
\newblock 2009.

\bibitem[Kurita et~al.(2020{\natexlab{a}})Kurita, Michel, and
  Neubig]{kurita-etal-2020-weight}
Keita Kurita, Paul Michel, and Graham Neubig.
\newblock Weight poisoning attacks on pretrained models.
\newblock In \emph{Proceedings of the 58th Annual Meeting of the Association
  for Computational Linguistics}, pp.\  2793--2806, Online, July
  2020{\natexlab{a}}. Association for Computational Linguistics.

\bibitem[Kurita et~al.(2020{\natexlab{b}})Kurita, Michel, and
  Neubig]{kurita2020weight}
Keita Kurita, Paul Michel, and Graham Neubig.
\newblock Weight poisoning attacks on pre-trained models.
\newblock \emph{arXiv preprint arXiv:2004.06660}, 2020{\natexlab{b}}.

\bibitem[Li et~al.(2020)Li, Xue, Zhao, Zhu, and Zhang]{li2020invisible}
Shaofeng Li, Minhui Xue, Benjamin Zhao, Haojin Zhu, and Xinpeng Zhang.
\newblock Invisible backdoor attacks on deep neural networks via steganography
  and regularization.
\newblock \emph{IEEE Transactions on Dependable and Secure Computing}, 2020.

\bibitem[Liao et~al.(2018)Liao, Zhong, Squicciarini, Zhu, and
  Miller]{liao2018backdoor}
Cong Liao, Haoti Zhong, Anna Squicciarini, Sencun Zhu, and David Miller.
\newblock Backdoor embedding in convolutional neural network models via
  invisible perturbation.
\newblock \emph{arXiv preprint arXiv:1808.10307}, 2018.

\bibitem[Lin et~al.(2020)Lin, Xu, Liu, and Zhang]{lin2020composite}
Junyu Lin, Lei Xu, Yingqi Liu, and Xiangyu Zhang.
\newblock Composite backdoor attack for deep neural network by mixing existing
  benign features.
\newblock In \emph{Proceedings of the 2020 ACM SIGSAC Conference on Computer
  and Communications Security}, pp.\  113--131, 2020.

\bibitem[Liu et~al.(2018)Liu, Dolan-Gavitt, and Garg]{liu2018fine}
Kang Liu, Brendan Dolan-Gavitt, and Siddharth Garg.
\newblock Fine-pruning: Defending against backdooring attacks on deep neural
  networks.
\newblock In \emph{International Symposium on Research in Attacks, Intrusions,
  and Defenses}, pp.\  273--294. Springer, 2018.

\bibitem[Liu et~al.(2017)Liu, Ma, Aafer, Lee, Zhai, Wang, and
  Zhang]{liu2017trojaning}
Yingqi Liu, Shiqing Ma, Yousra Aafer, Wen-Chuan Lee, Juan Zhai, Weihang Wang,
  and Xiangyu Zhang.
\newblock Trojaning attack on neural networks.
\newblock 2017.

\bibitem[Liu et~al.(2019)Liu, Lee, Tao, Ma, Aafer, and Zhang]{liu2019abs}
Yingqi Liu, Wen-Chuan Lee, Guanhong Tao, Shiqing Ma, Yousra Aafer, and Xiangyu
  Zhang.
\newblock Abs: Scanning neural networks for back-doors by artificial brain
  stimulation.
\newblock In \emph{Proceedings of the 2019 ACM SIGSAC Conference on Computer
  and Communications Security}, pp.\  1265--1282, 2019.

\bibitem[Liu et~al.(2020)Liu, Ma, Bailey, and Lu]{liu2020reflection}
Yunfei Liu, Xingjun Ma, James Bailey, and Feng Lu.
\newblock Reflection backdoor: A natural backdoor attack on deep neural
  networks.
\newblock In \emph{European Conference on Computer Vision}, pp.\  182--199.
  Springer, 2020.

\bibitem[Meng et~al.(2020)Meng, Fan, Sun, Hovy, Wu, and Li]{meng2020pair}
Yuxian Meng, Chun Fan, Zijun Sun, Eduard Hovy, Fei Wu, and Jiwei Li.
\newblock Pair the dots: Jointly examining training history and test stimuli
  for model interpretability.
\newblock \emph{arXiv preprint arXiv:2010.06943}, 2020.

\bibitem[Nguyen \& Tran(2020)Nguyen and Tran]{nguyen2020input}
Anh Nguyen and Anh Tran.
\newblock Input-aware dynamic backdoor attack.
\newblock \emph{arXiv preprint arXiv:2010.08138}, 2020.

\bibitem[Papineni et~al.(2002)Papineni, Roukos, Ward, and
  Zhu]{papineni2002bleu}
Kishore Papineni, Salim Roukos, Todd Ward, and Wei-Jing Zhu.
\newblock Bleu: a method for automatic evaluation of machine translation.
\newblock In \emph{Proceedings of the 40th annual meeting of the Association
  for Computational Linguistics}, pp.\  311--318, 2002.

\bibitem[Qi et~al.(2020)Qi, Chen, Li, Liu, and Sun]{qi2020onion}
Fanchao Qi, Yangyi Chen, Mukai Li, Zhiyuan Liu, and Maosong Sun.
\newblock Onion: A simple and effective defense against textual backdoor
  attacks.
\newblock \emph{arXiv preprint arXiv:2011.10369}, 2020.

\bibitem[Qi et~al.(2021{\natexlab{a}})Qi, Li, Chen, Zhang, Liu, Wang, and
  Sun]{qi2021hidden}
Fanchao Qi, Mukai Li, Yangyi Chen, Zhengyan Zhang, Zhiyuan Liu, Yasheng Wang,
  and Maosong Sun.
\newblock Hidden killer: Invisible textual backdoor attacks with syntactic
  trigger.
\newblock \emph{arXiv preprint arXiv:2105.12400}, 2021{\natexlab{a}}.

\bibitem[Qi et~al.(2021{\natexlab{b}})Qi, Yao, Xu, Liu, and Sun]{qi2021turn}
Fanchao Qi, Yuan Yao, Sophia Xu, Zhiyuan Liu, and Maosong Sun.
\newblock Turn the combination lock: Learnable textual backdoor attacks via
  word substitution.
\newblock \emph{arXiv preprint arXiv:2106.06361}, 2021{\natexlab{b}}.

\bibitem[Qiao et~al.(2019)Qiao, Yang, and Li]{qiao2019defending}
Ximing Qiao, Yukun Yang, and Hai Li.
\newblock Defending neural backdoors via generative distribution modeling.
\newblock \emph{arXiv preprint arXiv:1910.04749}, 2019.

\bibitem[Qiu et~al.(2021)Qiu, Zeng, Zheng, Guo, Zhang, and
  Li]{qiu2021efficient}
Han Qiu, Yi~Zeng, Qinkai Zheng, Shangwei Guo, Tianwei Zhang, and Hewu Li.
\newblock An efficient preprocessing-based approach to mitigate advanced
  adversarial attacks.
\newblock \emph{IEEE Transactions on Computers}, 2021.

\bibitem[Quiring \& Rieck(2020)Quiring and Rieck]{quiring2020backdooring}
Erwin Quiring and Konrad Rieck.
\newblock Backdooring and poisoning neural networks with image-scaling attacks.
\newblock In \emph{2020 IEEE Security and Privacy Workshops (SPW)}, pp.\
  41--47. IEEE, 2020.

\bibitem[Saha et~al.(2020)Saha, Subramanya, and Pirsiavash]{saha2020hidden}
Aniruddha Saha, Akshayvarun Subramanya, and Hamed Pirsiavash.
\newblock Hidden trigger backdoor attacks.
\newblock In \emph{Proceedings of the AAAI Conference on Artificial
  Intelligence}, volume~34, pp.\  11957--11965, 2020.

\bibitem[Salem et~al.(2020)Salem, Sautter, Backes, Humbert, and
  Zhang]{salem2020baaan}
Ahmed Salem, Yannick Sautter, Michael Backes, Mathias Humbert, and Yang Zhang.
\newblock Baaan: Backdoor attacks against autoencoder and gan-based machine
  learning models.
\newblock \emph{arXiv preprint arXiv:2010.03007}, 2020.

\bibitem[Socher et~al.(2013)Socher, Perelygin, Wu, Chuang, Manning, Ng, and
  Potts]{socher2013recursive}
Richard Socher, Alex Perelygin, Jean Wu, Jason Chuang, Christopher~D Manning,
  Andrew~Y Ng, and Christopher Potts.
\newblock Recursive deep models for semantic compositionality over a sentiment
  treebank.
\newblock In \emph{Proceedings of the 2013 conference on empirical methods in
  natural language processing}, pp.\  1631--1642, 2013.

\bibitem[Tran et~al.(2018)Tran, Li, and Madry]{tran2018spectral}
Brandon Tran, Jerry Li, and Aleksander Madry.
\newblock Spectral signatures in backdoor attacks.
\newblock \emph{arXiv preprint arXiv:1811.00636}, 2018.

\bibitem[Turner et~al.(2019)Turner, Tsipras, and Madry]{turner2019label}
Alexander Turner, Dimitris Tsipras, and Aleksander Madry.
\newblock Label-consistent backdoor attacks.
\newblock \emph{arXiv preprint arXiv:1912.02771}, 2019.

\bibitem[Vaswani et~al.(2017)Vaswani, Shazeer, Parmar, Uszkoreit, Jones, Gomez,
  Kaiser, and Polosukhin]{vaswani2017attention}
Ashish Vaswani, Noam Shazeer, Niki Parmar, Jakob Uszkoreit, Llion Jones,
  Aidan~N Gomez, {\L}ukasz Kaiser, and Illia Polosukhin.
\newblock Attention is all you need.
\newblock In \emph{Advances in neural information processing systems}, pp.\
  5998--6008, 2017.

\bibitem[Villarreal-Vasquez \& Bhargava(2020)Villarreal-Vasquez and
  Bhargava]{villarreal2020confoc}
Miguel Villarreal-Vasquez and Bharat Bhargava.
\newblock Confoc: Content-focus protection against trojan attacks on neural
  networks.
\newblock \emph{arXiv preprint arXiv:2007.00711}, 2020.

\bibitem[Wallace et~al.(2020)Wallace, Zhao, Feng, and
  Singh]{wallace2020concealed}
Eric Wallace, Tony~Z Zhao, Shi Feng, and Sameer Singh.
\newblock Concealed data poisoning attacks on nlp models.
\newblock \emph{arXiv preprint arXiv:2010.12563}, 2020.

\bibitem[Wang et~al.(2020{\natexlab{a}})Wang, Cao, Gong,
  et~al.]{wang2020certifying}
Binghui Wang, Xiaoyu Cao, Neil~Zhenqiang Gong, et~al.
\newblock On certifying robustness against backdoor attacks via randomized
  smoothing.
\newblock \emph{arXiv preprint arXiv:2002.11750}, 2020{\natexlab{a}}.

\bibitem[Wang et~al.(2019)Wang, Yao, Shan, Li, Viswanath, Zheng, and
  Zhao]{wang2019neural}
Bolun Wang, Yuanshun Yao, Shawn Shan, Huiying Li, Bimal Viswanath, Haitao
  Zheng, and Ben~Y Zhao.
\newblock Neural cleanse: Identifying and mitigating backdoor attacks in neural
  networks.
\newblock In \emph{2019 IEEE Symposium on Security and Privacy (SP)}, pp.\
  707--723. IEEE, 2019.

\bibitem[Wang et~al.(2020{\natexlab{b}})Wang, Nepal, Rudolph, Grobler, Chen,
  and Chen]{wang2020backdoor}
Shuo Wang, Surya Nepal, Carsten Rudolph, Marthie Grobler, Shangyu Chen, and
  Tianle Chen.
\newblock Backdoor attacks against transfer learning with pre-trained deep
  learning models.
\newblock \emph{IEEE Transactions on Services Computing}, 2020{\natexlab{b}}.

\bibitem[Weber et~al.(2020)Weber, Xu, Karla{\v{s}}, Zhang, and
  Li]{weber2020rab}
Maurice Weber, Xiaojun Xu, Bojan Karla{\v{s}}, Ce~Zhang, and Bo~Li.
\newblock Rab: Provable robustness against backdoor attacks.
\newblock \emph{arXiv preprint arXiv:2003.08904}, 2020.

\bibitem[Yang et~al.(2021)Yang, Li, Zhang, Ren, Sun, and He]{yang2021careful}
Wenkai Yang, Lei Li, Zhiyuan Zhang, Xuancheng Ren, Xu~Sun, and Bin He.
\newblock Be careful about poisoned word embeddings: Exploring the
  vulnerability of the embedding layers in nlp models.
\newblock \emph{arXiv preprint arXiv:2103.15543}, 2021.

\bibitem[Zhai et~al.(2021)Zhai, Li, Zhang, Wu, Jiang, and
  Xia]{zhai2021backdoor}
Tongqing Zhai, Yiming Li, Ziqi Zhang, Baoyuan Wu, Yong Jiang, and Shu-Tao Xia.
\newblock Backdoor attack against speaker verification.
\newblock In \emph{ICASSP 2021-2021 IEEE International Conference on Acoustics,
  Speech and Signal Processing (ICASSP)}, pp.\  2560--2564. IEEE, 2021.

\bibitem[Zhang et~al.(2015)Zhang, Zhao, and LeCun]{Zhang2015CharacterlevelCN}
Xiang Zhang, Junbo~Jake Zhao, and Yann LeCun.
\newblock Character-level convolutional networks for text classification.
\newblock In \emph{NIPS}, 2015.

\bibitem[Zhang et~al.(2020)Zhang, Zhang, Ji, and Wang]{zhang2020trojaning}
Xinyang Zhang, Zheng Zhang, Shouling Ji, and Ting Wang.
\newblock Trojaning language models for fun and profit.
\newblock \emph{arXiv preprint arXiv:2008.00312}, 2020.

\bibitem[Zhao et~al.(2020)Zhao, Ma, Zheng, Bailey, Chen, and
  Jiang]{zhao2020clean}
Shihao Zhao, Xingjun Ma, Xiang Zheng, James Bailey, Jingjing Chen, and Yu-Gang
  Jiang.
\newblock Clean-label backdoor attacks on video recognition models.
\newblock In \emph{Proceedings of the IEEE/CVF Conference on Computer Vision
  and Pattern Recognition}, pp.\  14443--14452, 2020.

\bibitem[Zhu et~al.(2020)Zhu, Ning, Wang, Xin, and Wu]{zhu2020gangsweep}
Liuwan Zhu, Rui Ning, Cong Wang, Chunsheng Xin, and Hongyi Wu.
\newblock Gangsweep: Sweep out neural backdoors by gan.
\newblock In \emph{Proceedings of the 28th ACM International Conference on
  Multimedia}, pp.\  3173--3181, 2020.

\end{thebibliography}
\bibliographystyle{iclr2022_conference}

\newpage

\appendix
\section{Derivation of Influence Functions}
\label{sec:derivation}
In this section, we give the derivation of influence functions (Equation \ref{eq:influence}) from \citet{koh2017understanding}.

Let $\mathcal{R}(\bm{\theta})=\frac{1}{N}\sum_{i=1}^N\mathcal{L}(\bm{z}_i;\bm{\theta})$ be the training objective, and $\bm{\theta}_{\bm{z},\epsilon}$ minimizes the weighted version of $\mathcal{R}(\bm{\theta})$:
\begin{equation}
    \mathcal{R}(\bm{\theta}_{\bm{z},\epsilon})+\epsilon\mathcal{L}(\bm{z};\bm{\theta}_{\bm{z},\epsilon})
\end{equation}
Taking derivative for both expressions and set them to zero, we have:
\begin{equation}
    \begin{aligned}
        \nabla \mathcal{R}(\bm{\theta})&=0\\
        \nabla \mathcal{R}(\bm{\theta}_{\bm{z},\epsilon})+\nabla\epsilon\mathcal{L}(\bm{z};\bm{\theta}_{\bm{z},\epsilon})&=0
    \end{aligned}
\end{equation}
Applying first-order Taylor expansion centered at $\bm{\theta}$ to the second equation leads to:
\begin{equation}
\begin{aligned}
    &\left(\nabla \mathcal{R}(\bm{\theta})+\epsilon\nabla\mathcal{L}(\bm{z};\bm{\theta})\right)\\
    &\left(\nabla^2\mathcal{R}(\bm{\theta})+\epsilon\nabla^2\mathcal{L}(\bm{z};\bm{\theta})\right)\bm{\Delta}_\epsilon\approx 0
\end{aligned}
\end{equation}
where $\bm{\Delta}_\epsilon\triangleq\bm{\theta}_{\bm{z},\epsilon}-\bm{\theta}$. Then solve for $\bm{\Delta}_\epsilon$:
\begin{equation}
\begin{aligned}
    \bm{\Delta}_\epsilon\approx& -\left(\nabla^2\mathcal{R}(\bm{\theta})+\epsilon\nabla^2\mathcal{L}(\bm{z};\bm{\theta})\right)^{-1}\\
    &\left(\nabla \mathcal{R}(\bm{\theta})+\epsilon\nabla\mathcal{L}(\bm{z};\bm{\theta})\right)
\end{aligned}
\end{equation}
The next step is to drop terms with higher order of $\epsilon$ $(\ge 2)$ and plug in $\nabla \mathcal{R}(\bm{\theta})=0$:
\begin{equation}
    \bm{\Delta}_\epsilon\approx-\nabla^2\mathcal{R}(\bm{\theta})^{-1}\nabla\mathcal{L}(\bm{z};\bm{\theta})\epsilon
\end{equation}
which then finally gives the result of Equation \ref{eq:influence} by taking derivative with respect to $\epsilon$:
\begin{equation}
    \frac{\mathrm{d}\bm{\theta}_{\bm{z},\epsilon}}{\mathrm{d}\epsilon}\Big|_{\epsilon=0}=\frac{\mathrm{d}\bm{\Delta}_\epsilon}{\mathrm{d}\epsilon}\Big|_{\epsilon=0}=-\bm{H}_{\bm{\theta}}^{-1}\nabla_{\bm{\theta}}\mathcal{L}(\bm{z};\bm{\theta})
\end{equation}

\section{Maximum Average Sub-graph Search}
\label{sec:search}
The greedy search and agglomerative search methods are present at Algorithm \ref{alg:greedy} and Algorithm \ref{alg:agglomerative}.

\begin{algorithm}
        \DontPrintSemicolon
              \KwInput{Influence graph $\mathcal{G}=(\mathcal{V},\mathcal{E})$, the number of poisoned training examples $\varepsilon N$}
              \KwOutput{The maximum average sub-graph $\hat{\mathcal{G}}$}
              Initialize $\hat{\mathcal{V}}\gets\varnothing,\hat{\mathcal{E}}\gets\varnothing$\\
              Find the edge with the largest influence score $(n_i,n_j)$ along with its associated nodes $\{n_i,n_j\}$\\
              $\hat{\mathcal{V}}\gets\hat{\mathcal{V}}\cup\{n_i,n_j\},\hat{\mathcal{E}}\gets\hat{\mathcal{E}}\cup \{(n_i,n_j)\}$\\
               \While{$|\hat{\mathcal{V}}|<\varepsilon N$}
               {
                   bestNode $n^*$, bestScore $s^*\gets -\infty$\\
                    \For{$n\in\mathcal{V}\backslash\hat{\mathcal{V}}$}
                    {
                        score $s\gets \sum_{n_k\in\hat{\mathcal{V}}}\texttt{EdgeWeight}(n,n_k)$\\
                        \If{$s>s^*$}
                        {
                            $s^*\gets s$\\
                            $n^*\gets n$
                        }
                    }
                    $\hat{\mathcal{E}}\gets\hat{\mathcal{E}}\cup \{(n^*,n_k)\}_{n_k\in\hat{\mathcal{V}}},\hat{\mathcal{V}}\gets\hat{\mathcal{V}}\cup\{n^*\}$
               }
               \Return $\hat{\mathcal{G}}$
            \caption{Greedy search}
            \label{alg:greedy}
\end{algorithm}
        
\begin{algorithm}
            \KwInput{Influence graph $\mathcal{G}=(\mathcal{V},\mathcal{E})$, the number of poisoned training examples $\varepsilon N$}
            \KwOutput{The maximum average sub-graph $\hat{\mathcal{G}}$}
              Initialize each node $n_i$ as an independent sub-graph and arrange them into a sub-graph set $\mathbb{G}=\{\tilde{\mathcal{G}}_i\}_{i=1}^N$\\
              \While{$|\tilde{\mathcal{G}}|<\varepsilon N,\forall \tilde{\mathcal{G}}\in\mathbb{G}$}
              {
                mergedGraph $\tilde{\mathcal{G}}^*,\tilde{\mathcal{G}}^{**}$, bestScore $s^*\gets-\infty$\\
                \For{$\tilde{\mathcal{G}}_i,\tilde{\mathcal{G}}_j\in\mathbb{G}$}
                    {
                        score $s\gets\frac{1}{|\tilde{\mathcal{G}}_1\cup\tilde{\mathcal{G}}_2|}\sum_{(n_p,n_q)\in\tilde{\mathcal{G}}_1\cup\tilde{\mathcal{G}}_2}\texttt{EdgeWeight}(n_p,n_q)$
                        \If{$s>s^*$}
                        {
                            $s^*\gets s$\\
                            $\tilde{\mathcal{G}}^*,\tilde{\mathcal{G}}^{**}\gets\tilde{\mathcal{G}}_1,\tilde{\mathcal{G}}_2$
                        }
                    }
                    $\mathbb{G}\gets\mathbb{G}\cup\{\tilde{\mathcal{G}}^*\cup\tilde{\mathcal{G}}^{**}\},\mathbb{G}\gets(\mathbb{G}\backslash\tilde{\mathcal{G}}^*)\backslash\tilde{\mathcal{G}}^{**}$
              }
              $\hat{\mathcal{G}}\gets\tilde{\mathcal{G}}\in\mathbb{G}~\text{s.t.}~|\tilde{\mathcal{G}}|\ge\varepsilon N$\\
              \Return $\hat{\mathcal{G}}$
            \caption{Agglomerative search}
            \label{alg:agglomerative}
\end{algorithm}

\end{document}